\documentclass[letterpaper,10pt,conference]{ieeeconf}

\IEEEoverridecommandlockouts
\usepackage[utf8]{inputenc}
\usepackage[caption=false,font=footnotesize]{subfig}
\usepackage{graphicx}
\usepackage{amsmath,amssymb}
\usepackage{cite}
\usepackage{url}
\usepackage{float}
\usepackage{xcolor}

\begin{document}

\title{\LARGE \bf
Reinforcement-Learning-Based Assistance Reduces Squat Effort
with a Modular Hip--Knee Exoskeleton
}

\author{
Neethan Ratnakumar$^{1}$,
Mariya Huzaifa Tohfafarosh$^{1}$, Saanya Jauhri$^{1}$,
and Xianlian Zhou$^{1}$
\thanks{$^{1}$Department of Biomedical Engineering,
New Jersey Institute of Technology, Newark, NJ, USA.}%
\thanks{Corresponding author: Xianlian Zhou
({\tt\small alexzhou@njit.edu}).}%
}

\maketitle
\thispagestyle{empty}
\pagestyle{empty}

\begin{abstract}
Squatting is one of the most demanding lower-limb movements, requiring substantial muscular effort and coordination. Reducing the physical demands of this task through intelligent and personalized assistance has significant implications, particularly in
industries involving repetitive low-level assembly activities. In this study, we evaluated the effectiveness of a neural network controller for a modular Hip-Knee exoskeleton designed to assist squatting tasks. The neural network controller was trained via reinforcement learning (RL)
in a physics-based, human-exoskeleton interaction simulation environment. The controller generated real-time hip and knee assistance torques based on recent
joint-angle and velocity histories. Five healthy adults performed three-minute metronome-guided
squats under three conditions:  (1) no exoskeleton (No-Exo), (2) exoskeleton with Zero-Torque, and
(3) exoskeleton with active assistance (Assistance). Physiological effort was assessed using indirect
calorimetry and heart rate monitoring, alongside concurrent kinematic data
collection. Results show that the RL-based controller adapts to individuals by producing torque profiles tailored to each subject’s kinematics and timing.
Compared with the Zero-Torque and No-Exo condition, active assistance reduced the net metabolic rate by approximately 10\%, with minor reductions observed in heart rate. However, assisted trials also exhibited reduced squat depth, reflected by smaller hip and knee flexion. These preliminary findings suggest that the proposed controller can effectively lower physiological effort during repetitive squatting, motivating further improvements in both hardware design and control strategies. 
\end{abstract}


\section{Introduction}\label{sec:introduction}
Squatting is a high-demand lower-limb movement frequently performed in assembly, and industrial settings, and it is associated with elevated musculoskeletal risk, particularly when combined with lifting tasks, or high repetition rates\cite{jensen2008knee,amin2008occupation}. Such exposure results in substantial mechanical loading of the knee and hip joints, contributing to muscular fatigue and increasing the likelihood of overuse injuries and chronic musculoskeletal disorders \cite{escamilla2001knee,amin2008occupation}. To address the physical demands associated with repetitive squatting tasks, lower-limb exoskeletons have increasingly been investigated as assistive devices in occupational and industrial contexts, with the goal of mitigating work-related musculoskeletal disorders (WMSDs) \cite{wei2020hip}. While several studies have reported reductions in knee \cite{dai2025squat} and hip joint loading, lower-limb muscular effort \cite{wei2020hip,yu2019design,ranaweera2018development}, together with lower metabolic energy requirements \cite{kantharaju2022reducing,yu2025controlling}, during assisted squatting tasks \cite{wang2021semi, kantharaju2022reducing}, many existing systems still rely on traditional control methods such as fixed or predefined assistance strategies \cite{li2025research}. However, these approaches may be limited in their ability to accommodate individual variability, as workers differ in anthropometry, movement patterns, and squat descent and ascent velocities. Moreover, during squatting, motion at one joint influences loading at adjacent joints, yet most contemporary exoskeletons and control strategies treat the knee and hip joints independently. 

To overcome these limitations, recent work has explored learning-based control strategies for lower-limb assistance. For example, predictive human–exoskeleton simulations have been used to optimize knee exoskeleton assistance during lifting tasks, followed by supervised neural network training to map anthropometric parameters and lifting postures to assistive torque profiles, resulting in reduced knee extensor muscle activation during experimental validation \cite{arefeen2024artificial}. More recently, adaptive control approaches based on reinforcement learning (RL) have been investigated for squat-related assistance tasks, enabling assistance policies to be learned through interaction with the human–exoskeleton system rather than relying on pre-optimized trajectories \cite{luo2021reinforcement}.
Building on these advances, this study introduces an RL trained neural network (NN) controller that generates real-time hip and knee joint torques by continuously observing joint states (i.e., joint angles and velocities). The proposed approach is evaluated using a custom hip-knee exoskeleton in an experimental study involving five healthy participants, and the resulting physiological and biomechanical benefits are reported.

\section{Methodology}\label{sec:methodology}

\subsection{Exoskeleton Hardware}

\begin{figure}[t]
  \centering
  \includegraphics[width=\linewidth]{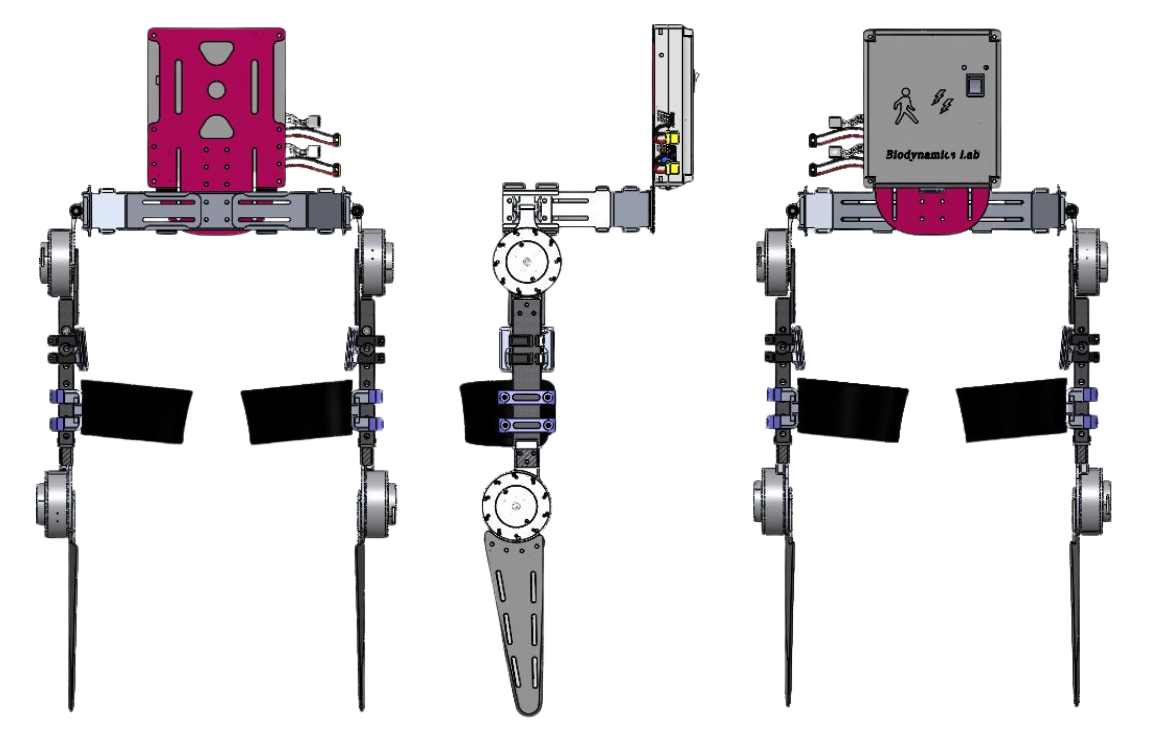}
  \caption{Modular hip--knee exoskeleton design showing front, side, and back views.}
  \label{fig:exo_cad}
\end{figure}

The custom Hip-Knee exoskeleton (Fig.~\ref{fig:exo_cad}) used in this study is a modular, backdrivable lower-limb device designed to provide medium-level torque support for movements such as squatting. It assists both hips and knees in the sagittal plane and weighs approximately 8.5~kgs. The structural frame consists of an aluminum waist assembly, a semirigid back plate, and telescopic carbon-fiber thigh links to accommodate different body sizes. A soft waist belt, padded shoulder straps, and adjustable thigh cuffs secure the device to the body while minimizing relative motion and maintaining comfort. The layout is modular and allows easy alignment of the actuator axes with the anatomical hip and knee joints.

Actuation is provided by four integrated motor modules (X8-25, MyActuator), each combining a pancake BLDC motor with an 1{:}9 planetary reduction, an embedded field-oriented control driver, and an 18-bit magnetic encoder; each delivers 10--13~Nm continuous and up to 25~Nm peak torque. In this study, assistance was limited to 10 Nm per joint.

A Raspberry Pi 4B executed the real-time NN controller and communicated with the actuators over CAN bus communication. Anatomical joint kinematics were estimated using inertial measurement units (Xsens MTw), since actuator encoders reflect motor rotation rather than true biological joint angles. A handheld safety-switch allowed immediate power cutoff during experiments. A custom PC-based graphical user interface (GUI)
was used for high-level experiment control, including mode selection, torque scaling,
and real-time visualization of joint kinematics and commanded torques via UDP
communication.

\subsection{Reinforcement Learning Training Framework for Squatting Assistance}

For the control of the exoskeleton, we developed a similar RL-based controller based on the learning in simulation framework \cite{luo2024experiment}.
To enable sim-to-real transfer, the RL training framework contains an Exoskeleton Control Network (ECN) that is
defined as an independent neural network interacting with two additional modules: the human
Control Policy Network (CPN) and the Muscle Coordination Network (MCN). The CPN
and MCN remain largely consistent with prior work\cite{ratnakumar2024predicting, luo2021reinforcement, luo2023robust, luo2024experiment}. The ECN independently receives joint-angle and angular-velocity
states and predicts normalized assistive torque outputs.

To simplify sim-to-real transfer and reduce dependence on device-specific parameters,
the human-exoskeleton interaction is modeled as idealized joint-torque assistance.
Because the exoskeleton joints are tightly coupled to the human joints, we assume that the
exoskeleton state closely matches the human joint state. This allows the ECN to use human
hip and knee joint angles and velocities as its input. This design has a practical advantage
for real-world implementation, since wearable sensors (e.g., IMUs) mounted directly on
the user can provide these measurements without relying on exoskeleton-side hardware.
This simplification reduces simulation complexity, accelerates training, and enables the ECN to be used across different exoskeleton platforms.

The ECN is implemented as a deterministic policy
\begin{equation}
a_e = \pi_{\psi_e}(s_e)
\label{eq:policy}
\end{equation}
where the network parameters $\psi_e$ are optimized via supervised learning. The network
consists of three fully connected hidden layers (64, 64, 64 nodes) with ReLU activations,
and a Tanh output layer that generates normalized assistance commands in the range
$[-1, 1]$. 
The supervised learning objective is

\begin{equation}
\begin{aligned}
\mathcal{L}(\psi)
&= \mathbb{E}\!\Big[
\left\| \boldsymbol{\tau}_{d} - \boldsymbol{\tau}_{exo} \right\|^{2}
+ w_{\text{reg}}
\left\| \boldsymbol{\tau}_{exo}(\psi) \right\|^{2} \\
&\quad
+ w_{\text{symm}}
\left\| \boldsymbol{\tau}_{exo}^L - \boldsymbol{\tau}_{exo}^R \right\|^{2}
\Big]
\end{aligned}
\label{eq:loss}
\end{equation}
where $\boldsymbol{\tau}_d$ denotes the desired human joint torques for squatting (normalized by the maximum assistance torque and clipped within [-1,1]), and
$\boldsymbol{\tau}_{exo}$ denotes the ECN predicted, normalized exoskeleton torques. The regularization term
($w_{\text{reg}} = 0.01$) discourages excessive torque magnitudes for improved comfort and
energy efficiency. 
The last term ($w_{\text{symm}} = 1.0$) is a bilateral symmetry loss term, reflecting the inherently symmetric nature of the squatting motion.

The RL component used to train the CPN employs a combination of rewards that
encourage accurate reproduction of reference squatting trajectories. A motion-matching
reward guides the simulated model toward a target squatting movement, while temporal
scaling of the reference allows the controller to learn across a range of squatting speeds.

Using this framework, an ECN was trained for the hip-knee exoskeleton to provide
real-time squatting assistance. The ECN receives a short history of hip and knee joint
angles and angular velocities at 100~Hz and outputs a normalized torque command. After
training, the ECN weights are exported and deployed on the exoskeleton control board
(Raspberry~Pi), where the final torque command is computed by scaling the ECN output
by a prescribed maximum torque (e.g., 10~Nm).

\begin{figure}[t]
  \centering
  \includegraphics[width=\linewidth]{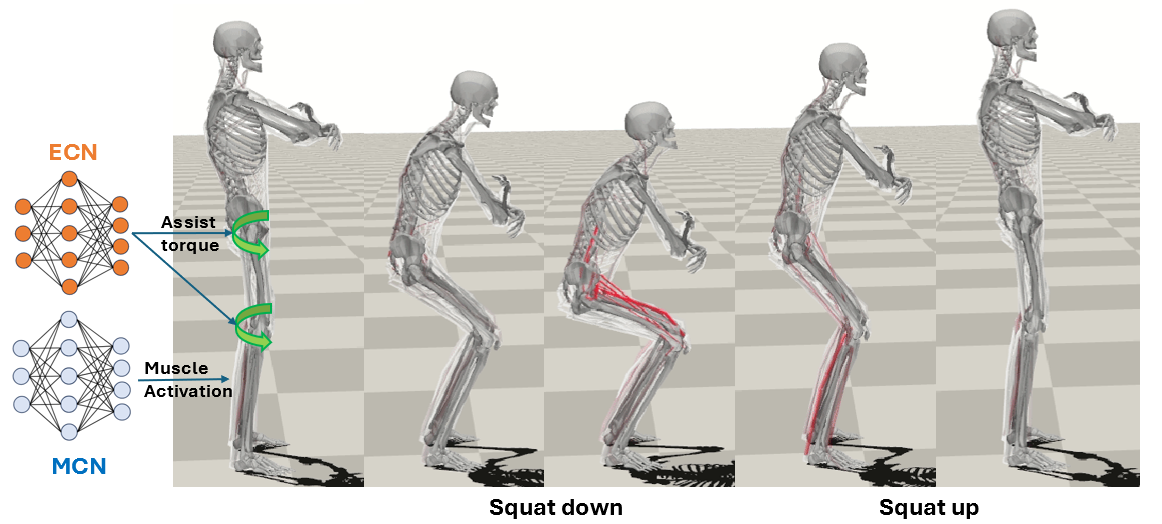}
  \caption{The RL learning framework with Exoskeleton Control Network (ECN) and Muscle Coordination Network (MCN). Idealized torque assistance (green curved arrows) is used instead of modeling the human-exoskeleton interaction. The human
Control Policy Network (CPN) is not presented here for brevity. The ECN predicts a normalized torque that is scaled by the maximum torque. The RL trained ECN and MCN work together to control the human model squatting up and down.}
  \label{fig:rl_sim}
\end{figure}

\subsubsection{Musculoskeletal model}\label{MSK-model}
The human body in our simulation is represented using a three-dimensional musculoskeletal model with 23 rigid body segments and 304 muscle–tendon actuators~\cite{park2022generative}. Each muscle connects to the skeleton through anatomical origin and insertion sites. Forces transmitted along the muscle path generate joint torques and contribute to limb movement. Muscle–tendon forces are modeled using a Hill-type formulation. In this framework, the total force produced by a musculotendon unit is composed of an activation-dependent contractile element together with a passive elastic contribution. For each muscle, the force is evaluated as
\begin{equation}
F(l,\dot{l},a)
=
F_0\left[\,a\,f_L(l)\,f_V(\dot{l}) + f_P(l)\,\right],
\label{eq:hill_model}
\end{equation}
where $F_0$ denotes the maximum isometric force under reference conditions, 
$l$ and $\dot{l}$ are the muscle fiber length and shortening velocity, and 
$a \in [0,1]$ is the activation level. 
The functions $f_L(l)$ and $f_V(\dot{l})$ represent the active force--length 
and force--velocity relationships, respectively, while 
$f_P(l)$ describes the passive force--length behavior. 
Active force therefore depends on both neural excitation and fiber mechanics, 
whereas passive force reflects the elastic tension that develops when the 
fiber is stretched beyond its slack length. 
The anatomical model supports subject-specific variation in body size and muscle properties~\cite{park2022generative}. Body segment dimensions can be scaled to represent different body shapes, while key muscle parameters can also be modified to capture clinically relevant changes in muscle function. 

\subsection{Experimental Protocol}\label{subsec:protocol}
Five healthy adults completed metronome-guided squats under three randomized conditions: (1) no exoskeleton (No-Exo), (2) exoskeleton with Zero-Torque, and (3) exoskeleton with active assistance (Assistance). Each condition lasted 3\,min at a standardized cadence set by the metronome. A brief familiarization session was conducted before data collection, and rest periods were provided between trials. Institutional approval and written informed consent were obtained. Safety measures included an emergency power cutoff to the motors.
Physiological and biomechanical signals were recorded during the squatting trials to evaluate exoskeleton effects. Lower-limb kinematics were captured with a wireless inertial motion system (Xsens Technologies B.V., Enschede, Netherlands) using sensors on the pelvis, thighs, shanks, and feet, sampled at 100\,Hz. Metabolic data were obtained via portable indirect calorimetry (K5, COSMED, Rome, Italy). Surface EMG was collected with a wireless system (Trigno, Delsys Inc., MA, USA), and heart rate was monitored using a chest-strap sensor (HRM-Dual, Garmin Ltd., KS, USA). Ground reaction forces were measured with a floor-mounted force plate (AMTI, MA, USA) during squatting. Plantar pressure data were acquired when participant shoe size permitted using insoles (OpenGo, Moticon ReGo AG, Munich, Germany).

\begin{figure}[t]
  \centering
  \includegraphics[width=0.925\linewidth]{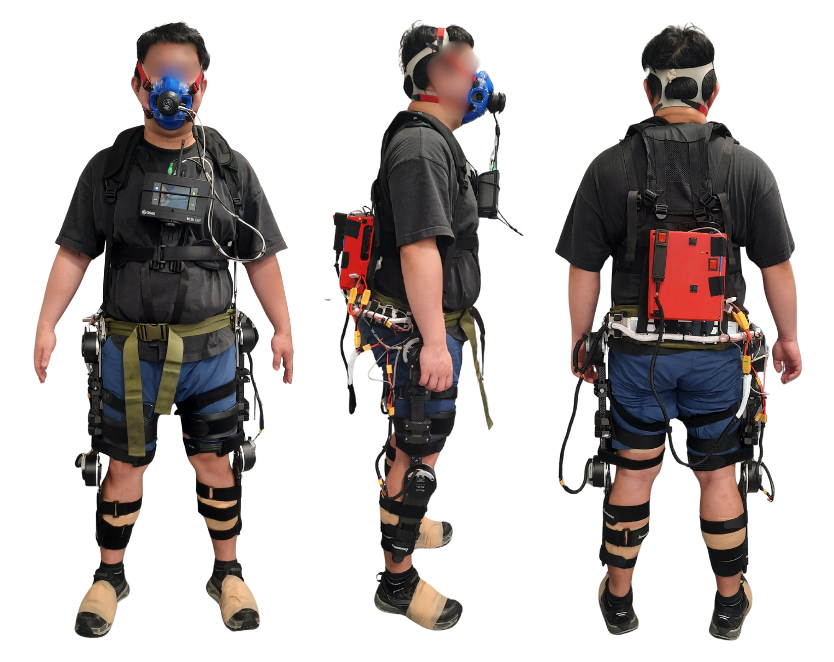}
\caption{A representative participant wearing the modular hip--knee exoskeleton
and metabolic measurement mask, shown from the front, side, and back.}
  \label{fig:exo_sub}
\end{figure}

\begin{figure}[t]
  \centering
  \includegraphics[width=0.92\linewidth]{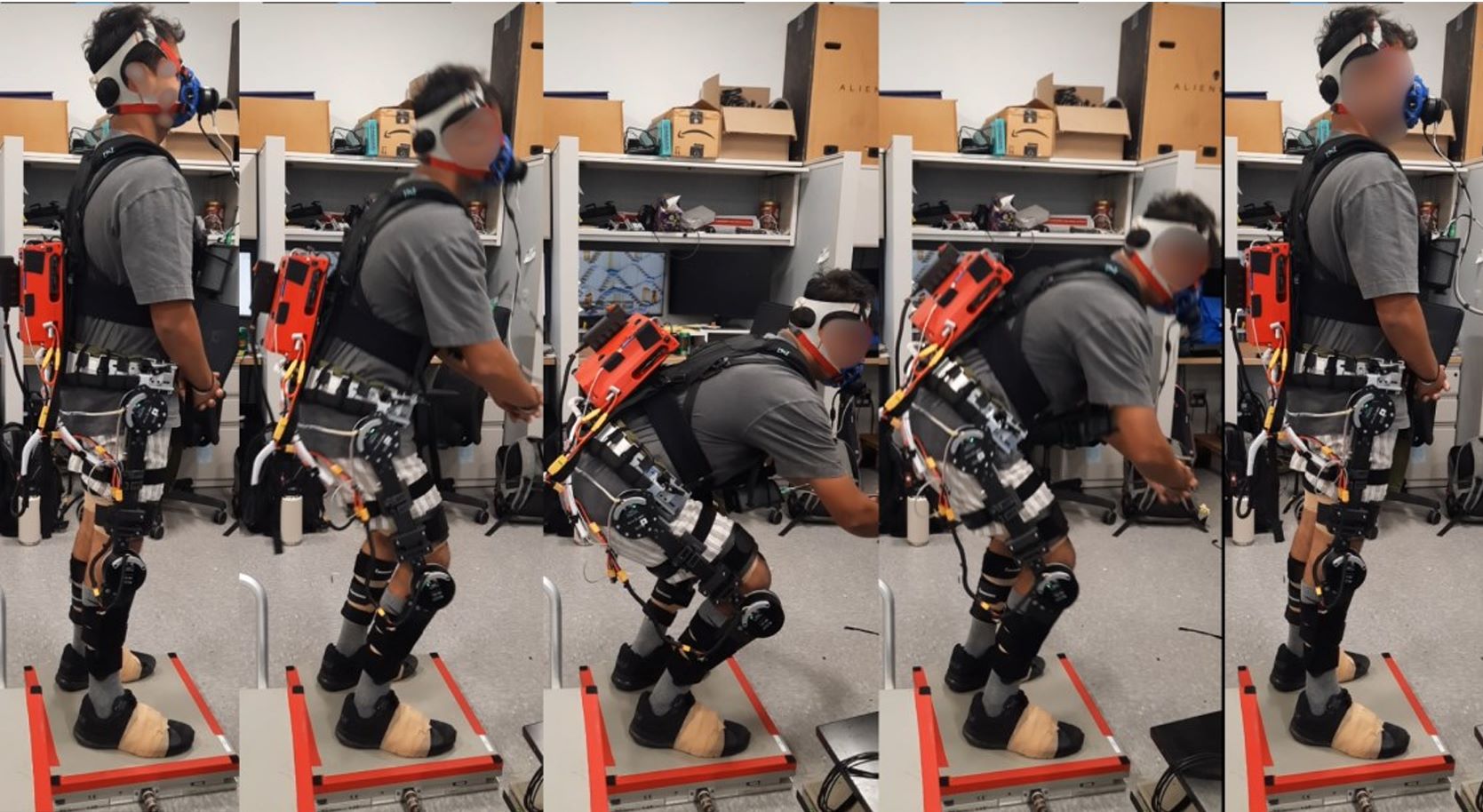}
\caption{Representative participant performing a squat cycle with the exoskeleton.}
  \label{fig:exo_squat}
\end{figure}

Experimental recording was coordinated through the custom GUI interface: EMG was controlled via TCP/IP to the Trigno Control Utility; Xsens recordings were triggered via UDP remote control; force plate capture (within Motive) was triggered from the same interface. The COSMED K5 received manual start/stop tags for temporal alignment. Exoskeleton logs (angles/velocities/commanded torques) were recorded concurrently, and all streams were aligned offline using triggers/tags and system timestamps. Representative still images of a subject performing the squatting task under 10 Nm peak exoskeleton assistance are shown in Fig.~\ref{fig:exo_squat}.

\subsection{Data Processing and Statistics}\label{subsec:stats}
Real-time collected data such as hip and knee joint angles (XSens streamed data), angular velocities (computed from angles via central difference), and NN commanded torques were stored during the experiment. 
Lower-limb kinematics data recorded by the Xsens IMU system were post-processed with the Reprocess HD engine using
the manufacturer’s software for high-quality joint angle data, which were exported for further offline analysis.
Oxygen uptake ($\dot{V}O_{2}$) and carbon dioxide production ($\dot{V}CO_{2}$) rates
(ml$\cdot$min$^{-1}$), measured using the COSMED K5 system, were converted to metabolic
power using the Brockway equation~\cite{brockway1987}. Resting standing metabolic
power was subtracted from gross values to obtain net metabolic power:

\begin{equation}
\dot{E} =
\frac{0.278\,\dot{V}O_{2} + 0.075\,\dot{V}CO_{2}}{m}
\label{eq:metrate}
\end{equation}

where $\dot{E}$ is metabolic power (W$\cdot$kg$^{-1}$) and $m$ is body mass (kg).


\section{Results}\label{sec:results}

\begin{figure}[ht]
  \centering
  \includegraphics[width=\linewidth]{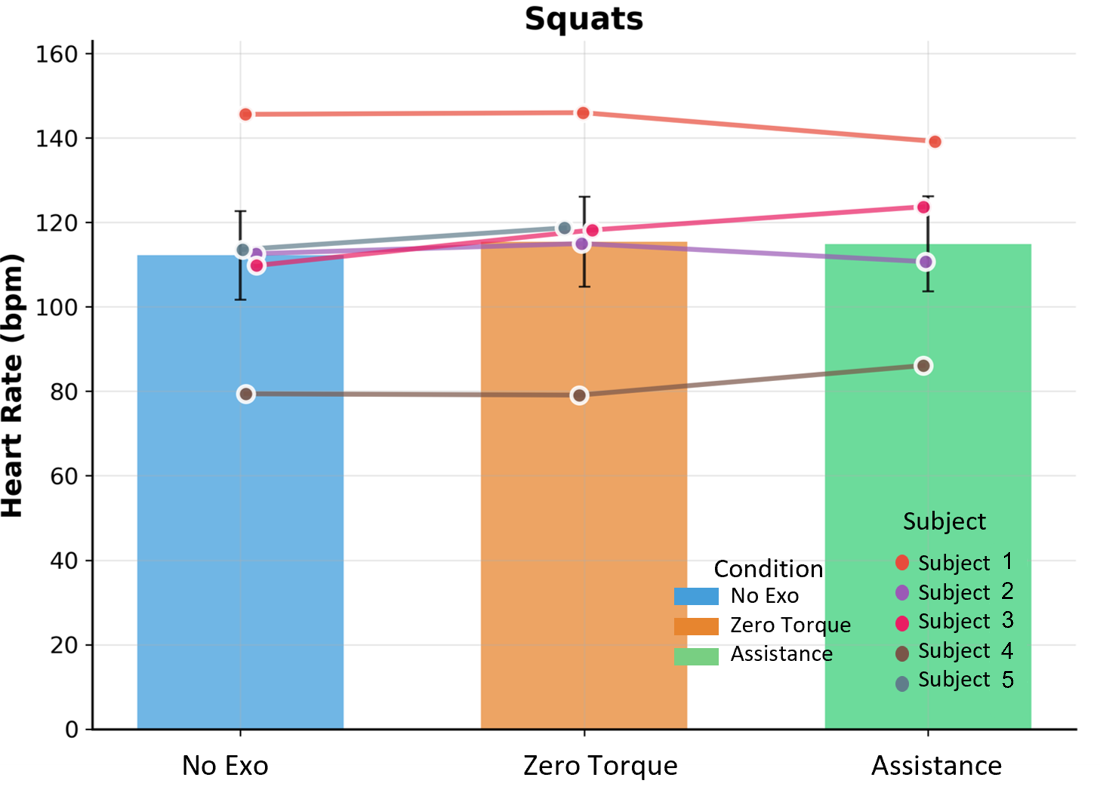}
  \caption{Heart Rate (bpm) by Condition}
  \label{fig:Net_hr}
\end{figure}

\begin{figure}[ht]
  \centering
  \includegraphics[width=\linewidth]{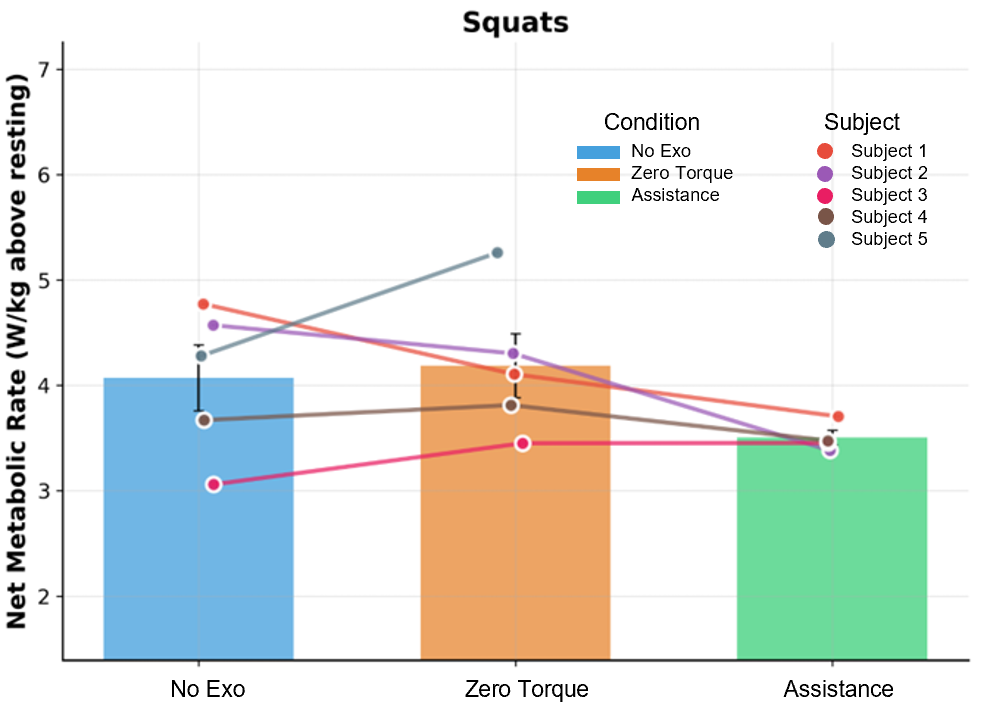}
  \caption{Net Metabolic Rate (W/kg above resting) by Condition}
  \label{fig:Net_met}
\end{figure}

On average, participants exhibited a resting (standing) metabolic rate of 1.71 W/kg and a resting heart rate of 91.8 bpm during the final three minutes of the baseline rest period. Fig. \ref{fig:Net_hr} shows the Heart Rate, and Fig. \ref{fig:Net_met} shows the Net Metabolic Rate (NMR) across conditions, while Table \ref{tab:hr_mr_squat} provides participant anthropometrics and computed percentage savings across conditions. The net metabolic rates in the Zero-Torque condition ranged from 3.451 to 5.259 W/kg, compared to 3.060 to 4.773 W/kg in the No-Exo condition. With active assistance, values ranged from 3.382 to 3.705 W/kg. Relative to the Zero-Torque condition, Assistance resulted in 8.8–21.4\% NMR reduction for three subjects and no reduction for the fourth subject. Across all subjects, the mean NMR reduction is 9.98\%. Relative to No Exo, Assistance produced changes of –0.39 to 1.19 W/kg (approximately –13\% to 26\%, with a mean of 10.2\%). Subject 5 did not complete the Assistance condition due to discomfort and fit issues and therefore was excluded from the Assistance statistics.

\begin{table*}[t]
\centering
\caption{Heart rate (HR) and net metabolic rate (NMR) during squatting under different exoskeleton conditions (Last 2 mins). Values are reported as mean of breath-to-breath measurements. Net metabolic rate (NMR) represents metabolic rate above resting baseline. In data rows, values in square brackets indicate percentage change relative to the Zero-Torque condition. Parentheses denote standard deviation across all subjects in the “Mean (SD)” row.}
\label{tab:hr_mr_squat}
\footnotesize
\setlength{\tabcolsep}{3pt}
\renewcommand{\arraystretch}{1.12}
\begin{tabular}{c c c c c c c c c}
\hline
Subject & Height & Weight &
\multicolumn{3}{c}{Heart Rate, HR (bpm)} &
\multicolumn{3}{c}{Net Metabolic Rate, NMR (W/kg)} \\
\cline{4-9}
 & (m) & (kg) & Zero-Torque & No-Exo& Assistance & Zero-Torque & No-Exo& Assistance \\
\hline
1 & 1.78 & 94    & 146.0 & 145.6\, [0.3] & 139.2\, [4.7] & 4.106 & 4.773\, [-16.3] & 3.705\, [9.8] \\
2 & 1.84 & 98.3  & 115.0 & 112.6\, [2.1] & 110.7\, [3.7] & 4.303 & 4.571\, [-6.2]  & 3.382\, [21.4] \\
3 & 1.80 & 102.3 & 118.2 & 109.8\, [7.1] & 123.7\, [-4.7] & 3.451 & 3.060\, [11.3]  & 3.453\, [-0.1] \\
4 & 1.52 & 73    & 79.1$^{*}$ & 79.4$^{*}$ & 86.1$^{*}$ & 3.813 & 3.670\, [3.8]   & 3.476\, [8.8] \\
5 & 1.70 & 64.65 & 118.7 & 113.6\, [4.3] & --$^{\dagger}$ & 5.259 & 4.279\, [18.6]  & --$^{\dagger}$ \\
\hline
\textbf{Mean (SD)} &
1.728 (0.127) & 86.45 (16.62) &
124.5 (14.4) & 120.4 (16.9) & 124.5 (14.3) &
4.186 (0.680) & 4.071 (0.702) & 3.504 (0.140) \\
\hline
\textbf{Mean \% change vs Zero-Torque} & -- & -- &
-- & 3.45 & 1.23 &
-- & 2.24 & 9.98 \\
\hline
\end{tabular}
\vspace{1mm}
\footnotesize

$^{*}$ HR data from Subject 4 were excluded from HR statistics due to invalid sensor readings.\\
$^{\dagger}$ Subject 5 did not complete the Assistance condition and was excluded from Assistance HR and MR statistics.
\end{table*}

\begin{figure}
    \centering
    \includegraphics[width=0.9\linewidth]{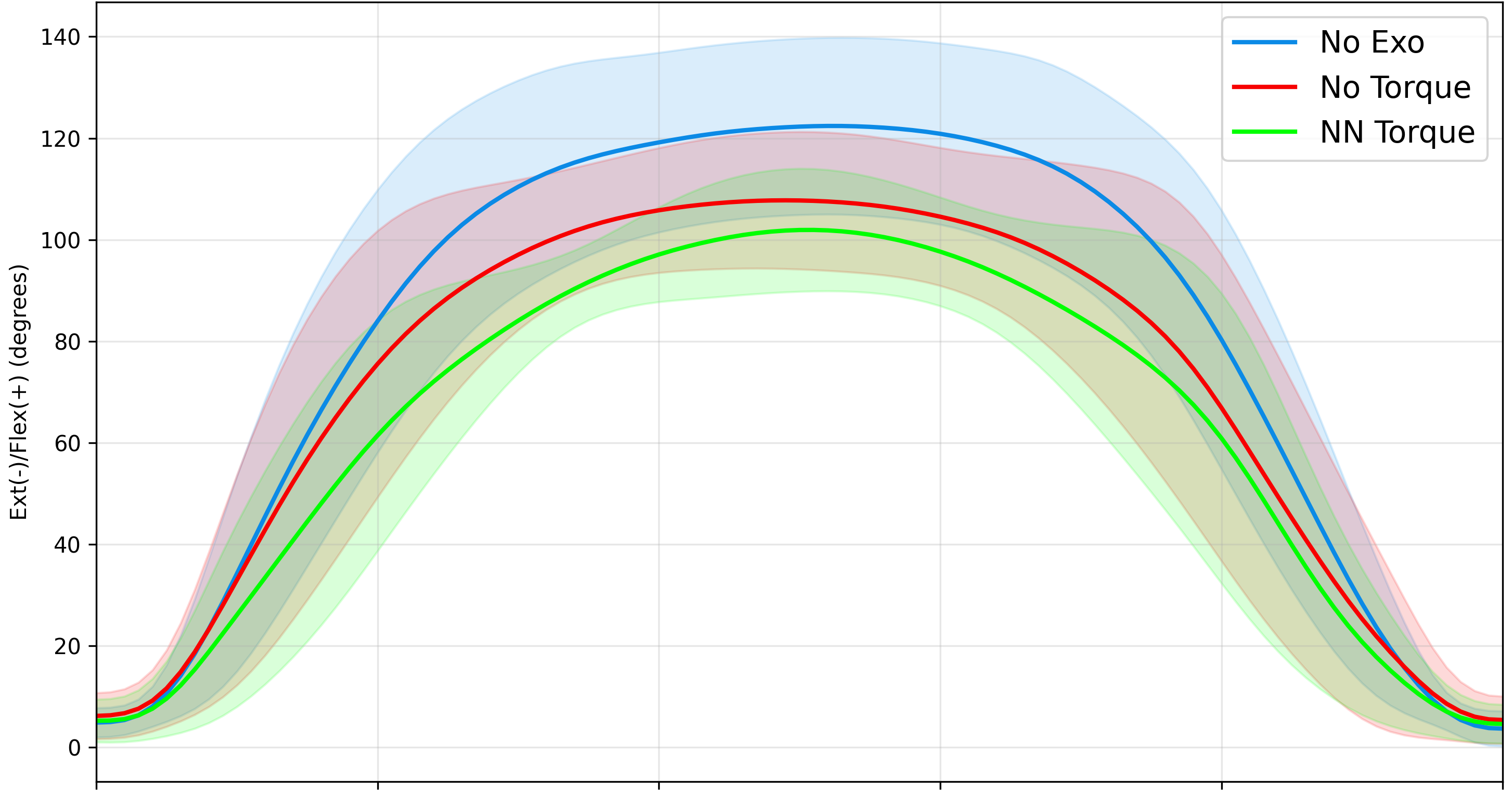}
    \caption{Left knee angle across the squat cycle (0-100\%) for all subjects under different conditions.}
    \label{fig:left_knee_angle_all_subjects}
\end{figure}

\begin{figure}
    \centering
    \includegraphics[width=0.9\linewidth]{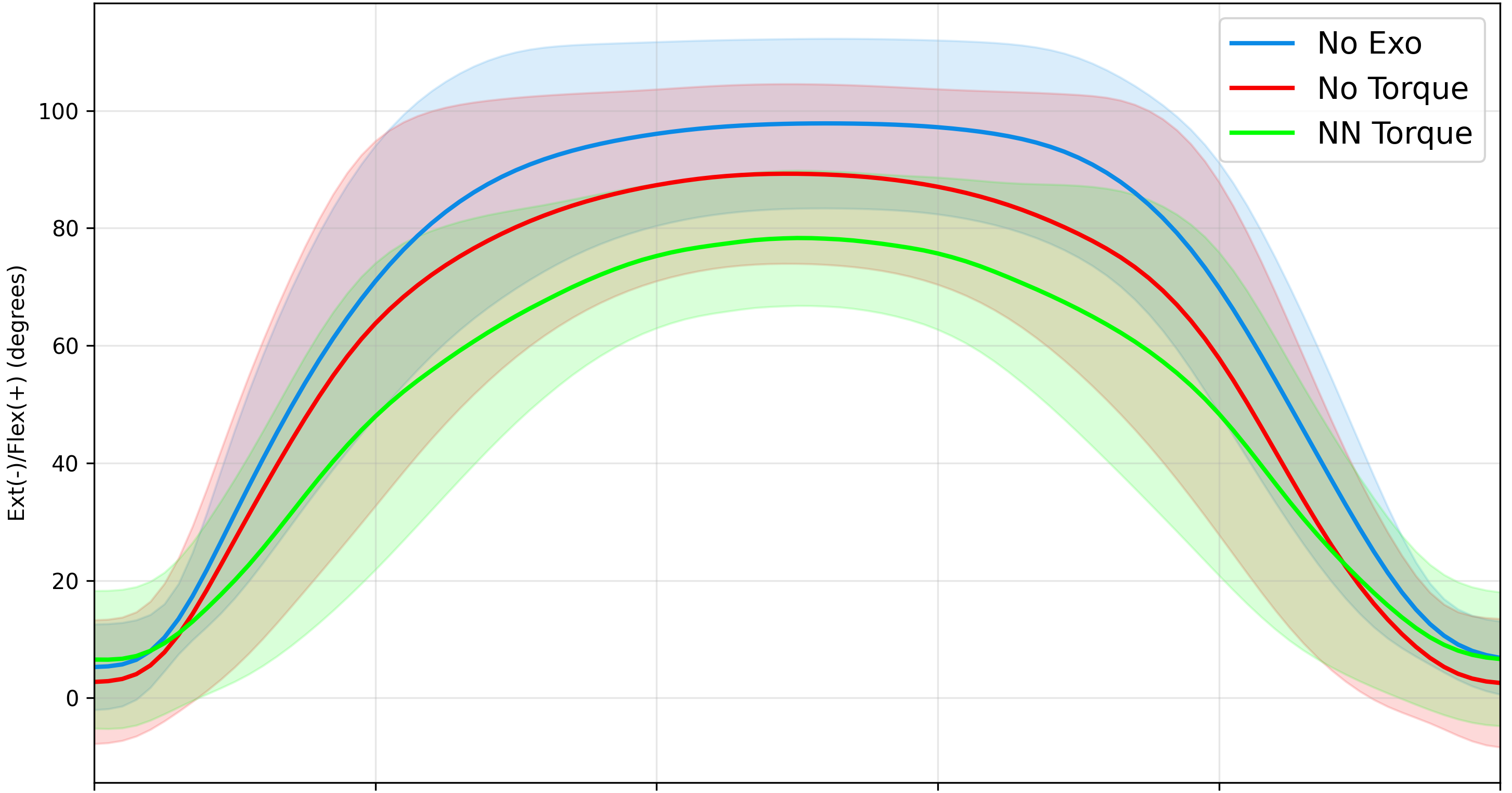}
    \caption{Left hip angle across the squat cycle (0-100\%) for all subjects under different conditions.}
    \label{fig:left_hip_angle_all_subjects}
\end{figure}

The lower-limb kinematics recorded through Xsens IMUs were HD processed, and their joint angles for the left knee and left hip joint for all conditions and all subjects are shown in Fig.~\ref{fig:left_knee_angle_all_subjects} and Fig.~\ref{fig:left_hip_angle_all_subjects}. The figures show the mean and standard deviation of the knee and hip angles, respectively, for all subjects combined across the No-Exo, Zero-Torque, and Assistance conditions. The peak knee and hip flexion angles were highest in the No-Exo condition, at 120 deg and 95 deg, respectively. Furthermore, the angle decreased when participants wore the exoskeleton. Peak knee and hip flexion were further reduced to 100 deg and 75 deg when assistance was applied. This decrement is primarily attributed to the mechanical constraints and kinematic impedance imposed by the device's structure. Furthermore, during the assistance mode, the active application of torque during both the descending and ascending phases, including the transition period, resulted in a more restricted range of motion compared to the unassisted conditions. This reduction in knee and hip flexion was consistently observed across all participants.

\begin{figure}[thpb]
\centering

\subfloat[Subject 3]{
  \includegraphics[width=\linewidth]{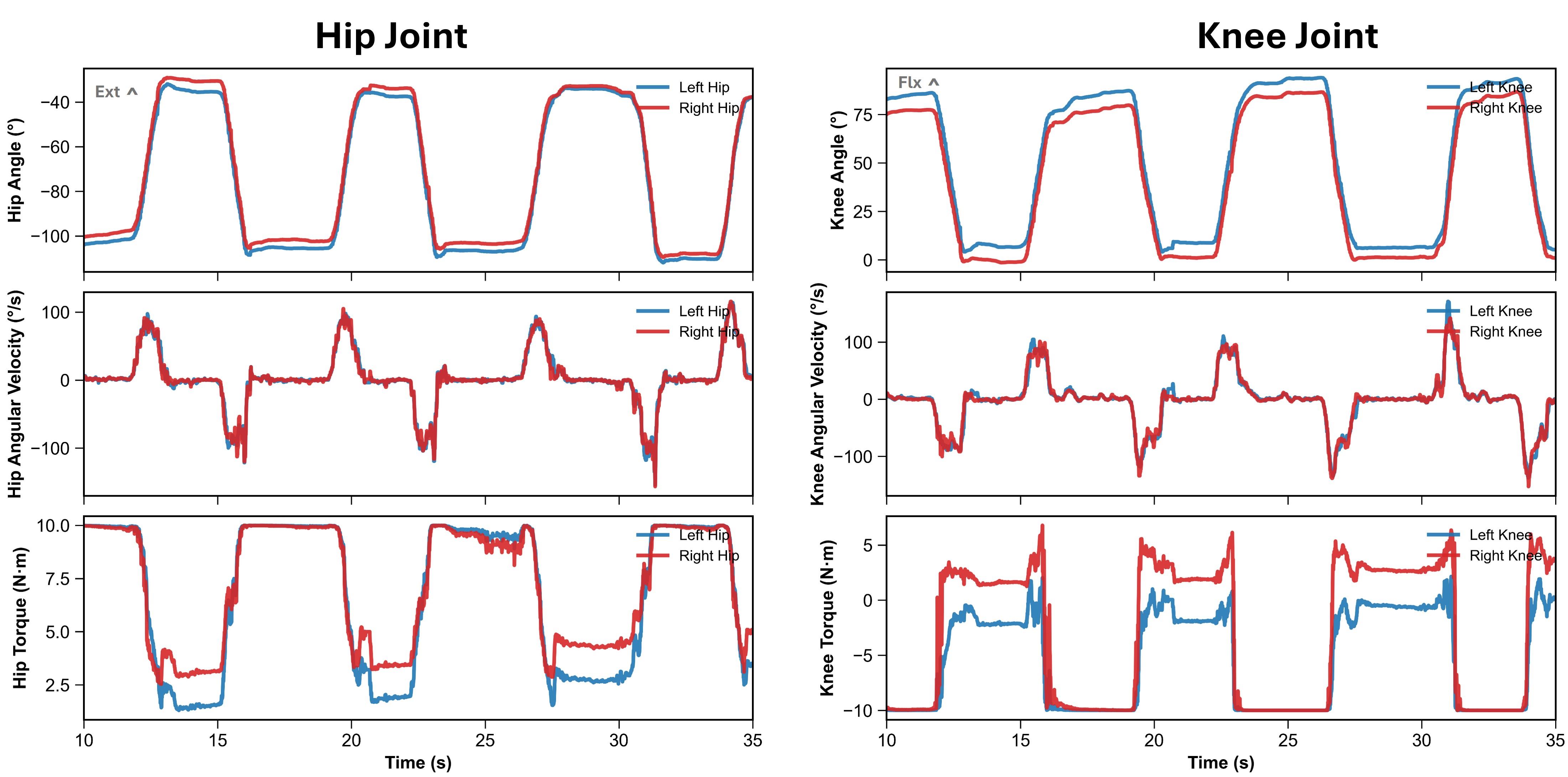}
  \label{fig:subject3}
}

\vspace{-4pt}

\subfloat[Subject 4]{
  \includegraphics[width=\linewidth]{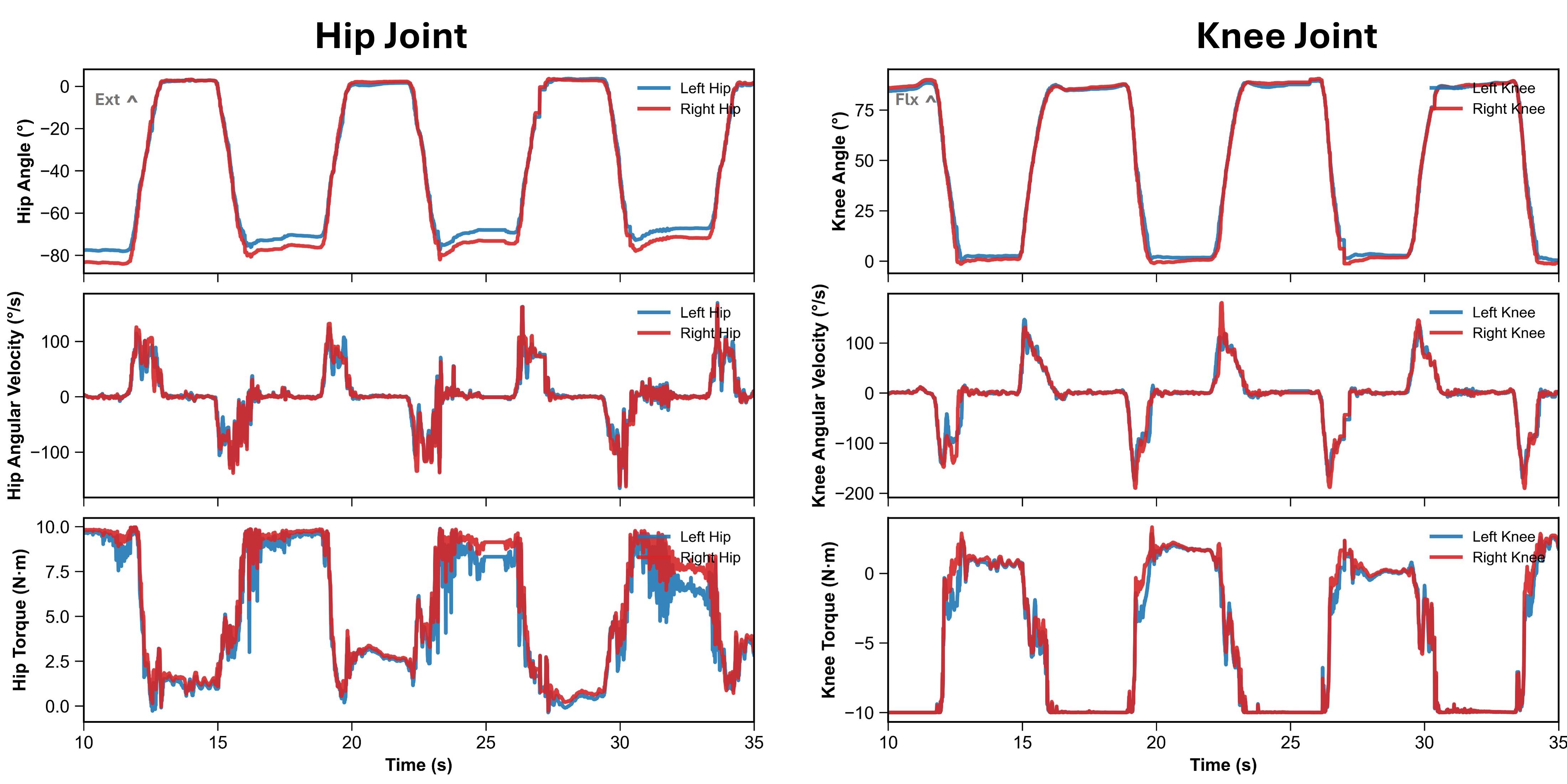}
  \label{fig:subject4}
}

\caption{Hip and knee angles, velocities, and applied torques during a portion of the assisted squat trials for two subjects. Positive values indicate extension for the hip plots, while positive values indicate flexion for the knee plots.}
\label{fig:Angle_Angular_Velocity_and_Torque}
\end{figure}

The angle, angular velocity, and commanded torque of the knee and hip for 2 subjects are depicted in the Fig. \ref{fig:Angle_Angular_Velocity_and_Torque}. The left side plots depict the hip joint, and the right side plots depict the knee joint. The angle and angular velocity histories were given as the input to the neural network, which calculates the commanded torque that is shown in the figure. As shown in Fig. \ref{fig:Angle_Angular_Velocity_and_Torque} (a), Subject 3 shows a hip angle offset of approximately 30 degrees while standing, potentially resulting from an offset in the Xsens IMU calibration. Additionally, the knee and hip torque for Subject 3 shows visible asymmetry between the applied torque on the left and right legs, which has been affected by the asymmetry in the joint angles.  In Fig. \ref{fig:Angle_Angular_Velocity_and_Torque} (b), for Subject 4, the applied hip torques show substantial jitter, which is directly inherited from the noise present in the noisy angular velocity signals. However, for Subject 4, the knee angular velocity and the applied torques look much smoother.

\section{Discussion}\label{sec:discussion}
This study examined the extent to which a simulation-informed, RL–based control strategy can reduce the physiological effort associated with repetitive squatting in healthy adults. Overall, the results demonstrate that real-time, adaptive hip–knee assistance can lower metabolic demand in the majority of participants, with reductions approaching 10\% on average and exceeding 20\% in the best-performing subject. These findings highlight the promise of learning-based controllers for personalized assistance during high-demand occupational movements such as squatting.

Inter-subject variability in metabolic response was observed, which reflects the inherent diversity in human movement strategies, adaptation rates, and device interaction dynamics. Importantly, the participant who demonstrated the largest benefits (Subject 2) had greater familiarity with the system, suggesting that user adaptation plays a key role in realizing the full potential of assistive control. This observation aligns well with a growing body of literature showing that repeated exposure and learning are critical for maximizing the benefits of wearable robotic assistance \cite{kim2019reducing,sawicki2008mechanics,wiggin2014characterizing,zhang2017human}. 

Several experimental factors may also have contributed to performance variability. Notably, this study was conducted as part of a broader multi-task protocol that included other activities. Prolonged testing periods can compromise measurement quality in several ways: participant fatigue and physical discomfort accumulate from prolonged device wear; calibration drifts may occur in the COSMED K5 system despite real-time corrections; and residual post-meal metabolic effects may influence baseline measurements \cite{westerterp2004diet}. Future work should therefore consider shorter testing blocks and baseline metabolic recalibration.

Hardware fit and sensing fidelity also emerged as important considerations. While the modular waist and thigh structures enabled rapid adjustment across participants, imperfect alignment for certain body types may have influenced both comfort and kinematic patterns. Likewise, the assistance policy relied on real-time joint angle and velocity estimates, making torque quality sensitive to sensing noise and communication latency. 
Nonetheless, these factors primarily represent engineering challenges rather than conceptual limitations of the control framework. Improvements in IMU filtering, synchronization, and mechanical compliance are likely to enhance torque smoothness and further stabilize assistance timing.





Assisted squatting was consistently associated with a reduction in peak hip and knee flexion angles. This kinematic shift suggests that participants adopted a modified movement strategy that effectively leveraged the mechanical support provided by the exoskeleton. Understanding how these strategy adaptations influence long-term musculoskeletal loading, comfort, and injury risk remains an important direction for future investigation.


\section{Conclusions}\label{sec:conclusions}
This study demonstrated the feasibility of deploying a RL-based neural-network controller for real-time assistance with a modular hip–knee exoskeleton during repetitive squatting. The proposed framework generated subject-specific torque profiles that adapted to individual movement timing and kinematics, resulting in reduced net metabolic cost for most participants. Assisted squatting was also associated with consistent changes in movement strategy, including reduced hip and knee flexion, suggesting that users naturally adapted to the provided assistance. While inter-subject variability was observed, the results highlight the potential of simulation-trained, learning-based controllers to deliver personalized assistance. Future work will focus on improving sensing robustness, hardware fit, and user adaptation to further enhance performance and consistency. In addition, analysis of sEMG will provide additional insight into the neuromuscular effects of exoskeleton assistance. With these advancements, the proposed framework is expected to support more uniform and robust personalized assistance for repetitive squatting.

\section*{Acknowledgment}
We sincerely thank all study participants for their time, commitment, and cooperation throughout the experiments. This work was partially supported by NSF award 2524089.
 

\end{document}